\documentclass[]{memtensor}
\usepackage{enumitem}
\usepackage[utf8]{inputenc}
\usepackage{url}
\usepackage{graphicx}
\usepackage{booktabs}
\usepackage{amsfonts}
\usepackage{nicefrac}
\usepackage{makecell}
\usepackage{microtype}
\usepackage{amsmath}
\usepackage{etoolbox}
\usepackage{lipsum}
\usepackage{minitoc}
\usepackage{tablefootnote}
\usepackage{threeparttable}
\usepackage{wrapfig}
\usepackage{appendix}
\usepackage{multirow}
\usepackage{ulem}
\useunder{\uline}{\ul}{}
\usepackage{colortbl}
\usepackage{adjustbox}
\usepackage{array}
\usepackage{tabularx}
\usepackage{siunitx}
\usepackage{placeins}
\usepackage{float}
\usepackage{xcolor}
\usepackage{mdframed}
\usepackage{listings}
\usepackage{amssymb}

\graphicspath{{img/}}
\IfFileExists{fontawesome5.sty}{
    \usepackage{fontawesome5}
}{
    
    \newcommand{\faGithub}{[GH]}
    \newcommand{\faPlug}{[API]}
    
}

\newcommand{\iconbox}[1]{\makebox[1.5em][c]{\large #1}}

\newcommand{\hficon}{\raisebox{-0.15em}{\IfFileExists{logo/huggingface.png}{\includegraphics[height=1.1em]{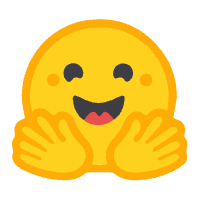}}{[HF]}}}

\newcommand{\figplaceholder}[1]{%
  \fbox{%
    \parbox[c][0.23\textheight][c]{0.9\linewidth}{\centering #1}%
  }%
}

\newcommand{\smartincludegraphics}[2]{%
  \IfFileExists{img/#1}{\includegraphics[width=#2]{#1}}{\figplaceholder{Missing figure: img/#1}}%
}

\lstdefinestyle{promptlisting}{
  basicstyle=\ttfamily\scriptsize,
  breaklines=true,
  breakatwhitespace=false,
  columns=fullflexible,
  keepspaces=true,
  showstringspaces=false,
  upquote=true,
  frame=none
}

\mdfdefinestyle{promptboxstyle}{
  linecolor=gray!60,
  linewidth=0.8pt,
  backgroundcolor=gray!8,
  roundcorner=2pt,
  innertopmargin=8pt,
  innerbottommargin=6pt,
  innerleftmargin=8pt,
  innerrightmargin=8pt,
  skipabove=8pt,
  skipbelow=8pt,
  frametitleaboveskip=0.5em,
  frametitlebelowskip=0.5em,
  frametitlerule=true,
  frametitlerulewidth=0.8pt,
  frametitlebackgroundcolor=gray!20,
  frametitlefont=\bfseries\small
}

\newenvironment{promptbox}[1]{%
  \begin{mdframed}[style=promptboxstyle,frametitle={#1}]
}{%
  \end{mdframed}
}

\title{%
  \titlefont%
  \parbox{\textwidth}{%
    \centering
    \raisebox{-0.3em}{\includegraphics[height=1.3em]{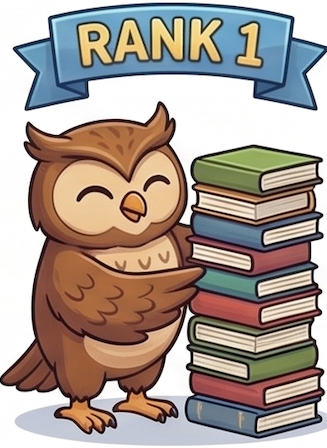}}\hspace{0.5em}%
    {\bfseries MemReranker:} Reasoning-Aware Reranking for Agent Memory Retrieval
  }%
}

\author[1]{Chunyu Li}
\author[1]{Mengyuan Zhang}
\author[1]{Jingyi Kang}
\author[2]{Ding Chen}
\author[1]{Jiajun Shen}
\author[1]{Bo Tang}
\author[3]{Xuanhe Zhou}
\author[1]{Feiyu Xiong}
\author[1]{Zhiyu Li\textsuperscript{ \faEnvelope,}}
 
\affiliation[1]{MemTensor (Shanghai) Technology} 
\affiliation[2]{China Telecom Research Institute}
\affiliation[3]{Shanghai Jiao Tong University}

\abstract{
In agent memory systems, the reranking model serves as the critical bridge connecting user queries with long-term memory. Most systems adopt the "retrieve-then-rerank" two-stage paradigm, but generic reranking models rely on semantic similarity matching and lack genuine reasoning capabilities, leading to a problem where recalled results are semantically highly relevant yet do not contain the key information needed to answer the question. This deficiency manifests in memory scenarios as three specific problems. First, relevance scores are miscalibrated, making threshold-based filtering difficult. Second, ranking degrades when facing temporal constraints, causal reasoning, and other complex queries. Third, the model cannot leverage dialogue context for semantic disambiguation. This report introduces MemReranker, a reranking model family (0.6B/4B) built on Qwen3-Reranker through multi-stage LLM knowledge distillation. Multi-teacher pairwise comparisons generate calibrated soft labels, BCE pointwise distillation establishes well-distributed scores, and InfoNCE contrastive learning enhances hard-sample discrimination. Training data combines general corpora with memory-specific multi-turn dialogue data covering temporal constraints, causal reasoning, and coreference resolution. On the memory retrieval benchmark, MemReranker-0.6B substantially outperforms BGE-Reranker and matches open-source 4B/8B models as well as GPT-4o-mini on key metrics. MemReranker-4B further achieves 0.737 MAP, with several metrics on par with Gemini-3-Flash, while maintaining inference latency at only 10--20\% of large models. On finance and healthcare vertical-domain benchmarks, the models preserve generalization capabilities on par with mainstream large-parameter rerankers.

\vspace{1.5em}
\noindent
\
\begingroup

\begin{tabular}{@{}l@{}}
\iconbox{\faGithub} \textbf{Github:} \href{https://github.com/MemTensor/MemOS}{github.com/MemTensor/MemOS} \\
\iconbox{\hficon} \textbf{MemReranker-4B:} \href{https://huggingface.co/IAAR-Shanghai/MemReranker-4B}{huggingface.co/IAAR-Shanghai/MemReranker-4B} \\
\iconbox{\faPlug} \textbf{API Documentation:} \href{https://memos-docs.openmem.net/cn/self_developed_model/reranker_usage_example}{docs.openmem.net/api} \\
\iconbox{\faEnvelope} \textbf{Correspondence:} \href{lizy@memtensor.cn}{lizy@memtensor.cn} \\
\end{tabular}
\endgroup
}

\begin{document}
\maketitle

\newpage

\section{Introduction}

\begin{figure*}[t]
    \centering
    \smartincludegraphics{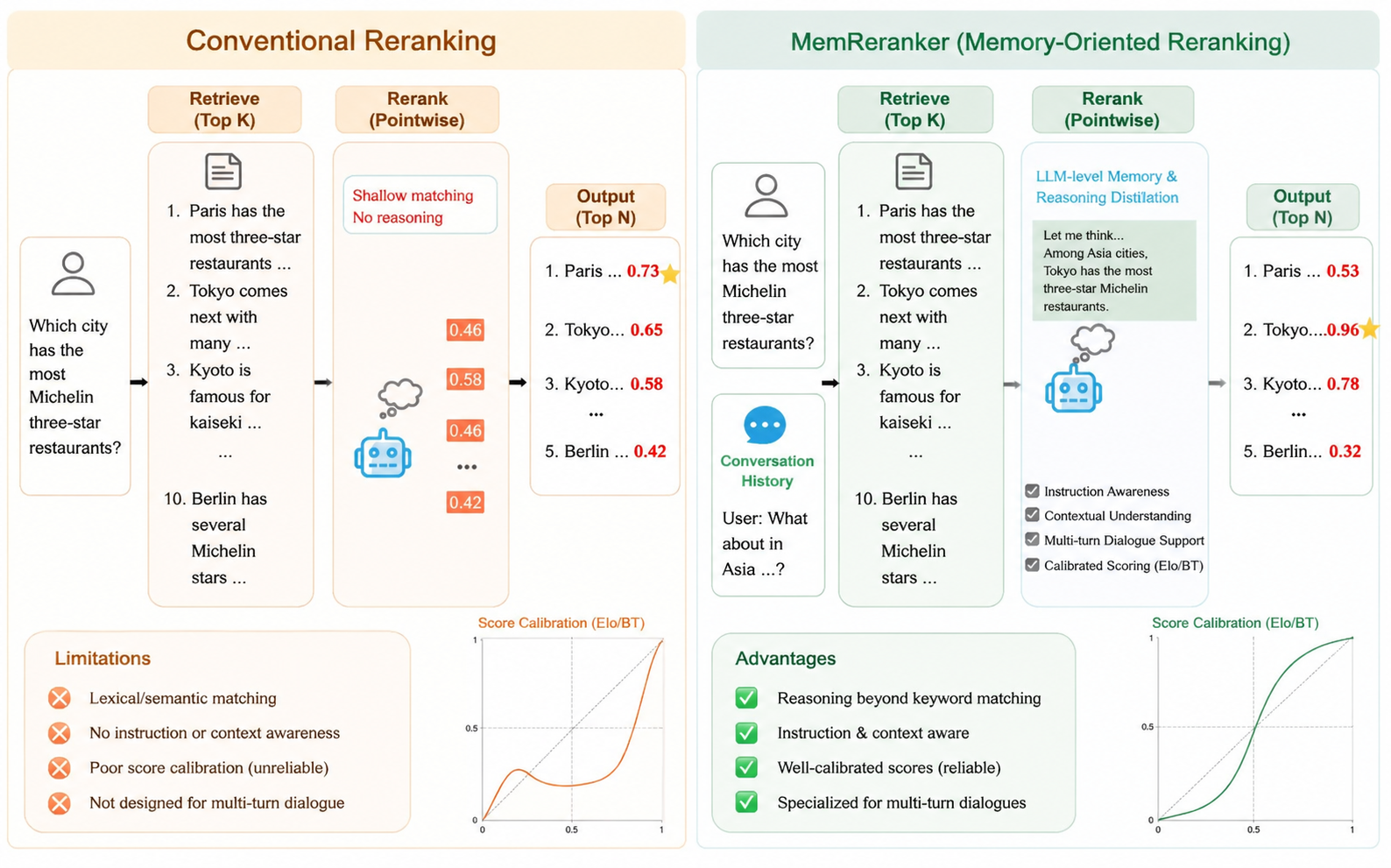}{0.98\textwidth}
    \caption{\textbf{Conventional reranking vs.\ MemReranker.}
    Left: generic rerankers rely on shallow semantic matching with no instruction or context awareness, producing poorly calibrated scores (left-skewed distribution).
    Right: MemReranker distills LLM-level reasoning into a compact model, incorporating instruction awareness, contextual understanding, multi-turn dialogue support, and Elo/Bradley-Terry calibrated scoring, yielding well-distributed relevance scores that enable reliable threshold-based filtering.}
    \label{fig:overview}
\end{figure*}

Long-term memory is becoming the core capability that transforms agent systems from single-turn tools into persistent companions~\cite{memorysurvey, memos}. Recent work has established memory as a first-class system resource through structured memory operating systems and explicit memory architectures~\cite{memory3}, while production-ready memory layers such as Mem0~\cite{mem0} and agentic memory frameworks like A-Mem~\cite{amem} demonstrate growing demand for memory-augmented agents. When an agent can accurately recall dietary preferences mentioned three months ago, project decisions discussed two weeks prior, or emotional states expressed yesterday, the interaction experience undergoes a qualitative leap. Realizing this capability depends on a seemingly simple yet extremely challenging technical component: precisely retrieving truly relevant memories from tens of thousands of historical dialogue fragments in response to a current query.

Existing systems generally adopt the ``retrieve-then-rerank'' two-stage paradigm, where dense vector models such as BGE-M3~\cite{bgem3} complete candidate recall and reranking models perform fine-grained ordering. However, this architecture harbors a widely underestimated fundamental problem: \textbf{generic reranking models rely heavily on semantic similarity for ranking, but semantic similarity does not equate to containing the answer.} In practical memory retrieval, a large number of dialogue fragments match the query highly on surface semantics yet do not contain the key information required to answer the question. The system returns a batch of seemingly relevant results, but the downstream large model cannot generate a correct answer. This problem has been systematically characterized by HaluMem~\cite{halumem}, which identifies memory hallucinations fabrication, errors, and omissions across extraction, updating, and retrieval stages. Meanwhile, MemRL~\cite{memrl} demonstrates that passive semantic matching in memory retrieval often retrieves noise, motivating value-based retrieval strategies. This fundamental deficiency further evolves into three specific problems in memory scenarios. \textbf{Score calibration failure:} The relevance scores of models such as BGE-Reranker exhibit extreme left-skewed distributions (most scores cluster near 0.00), making it virtually impossible for production systems to set effective cutoff thresholds. \textbf{Reasoning capability deficit:} Encoder-based rerankers rely on lexical and shallow semantic matching, performing poorly when facing queries requiring temporal constraints, numerical comparison, and causal logic reasoning. \textbf{Instruction awareness void}, Generic rerankers cannot leverage contextual instructions for semantic disambiguation the same query ``I want to look at Apple'' has entirely different meanings in a conversation about smartphones versus one about fruit, yet generic models cannot distinguish between them. Figure~\ref{fig:overview} illustrates this contrast, conventional rerankers produce left-skewed, poorly calibrated score distributions and lack reasoning beyond lexical matching, whereas MemReranker yields well separated scores through Elo/Bradley-Terry calibration while integrating instruction awareness, contextual understanding, and multi-turn dialogue support via LLM-level knowledge distillation.

In recent years, LLM-based reranking methods represented by RankGPT~\cite{rankgpt} and RankZephyr~\cite{rankzephyr} have demonstrated that large language models can effectively overcome these limitations through deep reasoning and world knowledge. However, deploying models with 7 billion or more parameters in real-time memory systems is infeasible in terms of both latency and cost. This raises the core question of this work: \textbf{Can we distill the reasoning capabilities of large-parameter LLM rerankers into compact models suitable for production deployment, with specialized enhancements for memory scenarios?}

To this end, we propose the MemReranker model family (0.6B / 4B) and design a progressive distillation training pipeline from label generation to contrastive learning, along with a specialized dataset constructed for multi-turn dialogue semantic shift. The main contributions of this paper are as follows:

\begin{itemize}
    \item We propose the MemReranker family of models for agent memory retrieval. With only 0.6B/4B parameters, they achieve ranking quality on par with closed-source large models GPT-4o-mini and Gemini-3-Flash on the LOCOMO benchmark, while reducing inference latency to approximately 200ms.
    \item We design a two-stage training paradigm combining BCE pointwise distillation and InfoNCE contrastive fine-tuning, integrated with a five-level calibrated scoring system based on Elo/Bradley-Terry, systematically addressing the score calibration and hard-sample discrimination challenges of small models.
    \item We construct a multi-turn dialogue data engineering pipeline tailored for memory scenarios, encompassing history distillation, hard negative generation, and instruction augmentation, enabling the model to learn coreference resolution and topic drift modeling from dialogue fragments.
    \item We conduct systematic evaluation and ablation analysis across memory retrieval, hard-case ranking, finance, and healthcare benchmarks, validating the effectiveness of each design choice.
\end{itemize}

\section{Related Work}

\subsection{LLM-Based Reranking}

The integration of large language models with document reranking has fundamentally reshaped the information retrieval landscape. RankGPT~\cite{rankgpt} pioneered the use of GPT-4 for zero-shot listwise reranking through carefully designed prompts. RankZephyr~\cite{rankzephyr} demonstrated that open-source 7B models, distilled from GPT-4 teacher rankings, can match or exceed proprietary model performance on the TREC Deep Learning track~\cite{trecdl} and BEIR~\cite{beir} benchmarks. The FIRST method~\cite{first} further improved efficiency by generating rankings from first-token logits rather than full sequence generation.

These advances established a clear hierarchy: LLM rerankers achieve superior quality through deep reasoning, but at prohibitively high computational cost (7B+ parameters, sequence generation overhead). This has spawned a wave of distillation research aimed at compressing LLM ranking knowledge into compact, deployable models.

\subsection{Knowledge Distillation for Reranking}

Table~\ref{tab:distillation} summarizes the key distillation methods that guided our methodology.

\begin{table}[H]
\centering
\small
\begin{adjustbox}{max width=\linewidth}
\begin{tabular}{lllll}
\toprule
\textbf{Method} & \textbf{Teacher} & \textbf{Student} & \textbf{Loss} & \textbf{Key Innovation} \\
\midrule
Rank-DistiLLM & RankZephyr 7B & MonoELECTRA 335M & Listwise + hard neg. & 173$\times$ faster than LLM reranker \\
DeAR & LLaMA2-13B & Qwen 1.7B--7B & CE + RankNet + KL $\to$ CoT & Two-stage: pointwise $\to$ listwise \\
zerank / zELO & LLM ensemble & Qwen3-1.7B & MSE & Pairwise $\to$ Elo $\to$ pointwise MSE \\
InRanker & monoT5-3B & T5-small 60M & MSE (soft labels) & 60M matches 3B teacher quality \\
BiXSE & LLM graded labels & Qwen2.5-0.5B & BCE (graded scores) & 1 label per query; BCE $>$ InfoNCE at 0.5B \\
\textbf{MemReranker} & GPT / Qwen ens. & Qwen3-0.6B/4B & BCE $\to$ InfoNCE (two-stage) & Memory-oriented; instruction-aware; multi-turn \\
\bottomrule
\end{tabular}
\end{adjustbox}
\caption{Summary of key distillation methods for reranking. MemReranker combines insights from multiple prior works and introduces memory-specific adaptations.}
\label{tab:distillation}
\end{table}

Rank-DistiLLM~\cite{rankdistillm} demonstrated that cross-encoders trained with LLM-distilled data can achieve LLM-level effectiveness while being 173$\times$ faster. DeAR~\cite{dear} proposed a two-stage pipeline---pointwise distillation followed by listwise chain-of-thought training---performing strongly even at 1.7B scale. zerank/zELO~\cite{zerank} contributed a principled score calibration approach by transforming pairwise LLM judgments into Elo/Thurstone scores, producing well-calibrated continuous outputs. InRanker~\cite{inranker} achieved the remarkable result of a 60M-parameter model matching its 3B teacher through soft-label MSE distillation. BiXSE~\cite{bixse} provided the key insight that at the 0.5B-parameter scale, BCE loss with LLM-graded scores outperforms InfoNCE contrastive learning---directly guiding our training strategy.

\subsection{Existing Reranker Architectures}

We surveyed the landscape of production reranking models to guide our architectural decisions. Table~\ref{tab:reranker_arch} presents the comparison.

\begin{table}[H]
\centering
\small
\begin{adjustbox}{max width=\linewidth}
\begin{tabular}{lllllll}
\toprule
\textbf{Model} & \textbf{Architecture} & \textbf{Output} & \textbf{Params} & \textbf{Max Len} & \textbf{Language} & \textbf{Key Feature} \\
\midrule
BGE-Reranker-v2-m3 & Encoder-only (XLM-R) & Logits (scalar) & 560M & 8192 & 100+ & Standard cross-encoder \\
BGE-Reranker-v2-Gemma & Decoder-only (Gemma) & Token prob. (Yes/No) & 2B / 9B & 8192 & English-centric & LLM-based cross-encoder \\
Qwen3-Reranker~\cite{qwen3embedding} & Decoder-only (Qwen3) & Gen. text / logits & 0.6B--8B & 32K+ & 100+ & Generative scoring + MRL \\
Jina-Reranker-v2 & Encoder-only (XLM-R) & Logits & 278M & 8192 & 89+ & Task adapters \\
RankZephyr & Decoder-only (Zephyr 7B) & Listwise permutation & 7B & 4096 & English-centric & GPT-4 distilled; zero-shot \\
\bottomrule
\end{tabular}
\end{adjustbox}
\caption{Overview of production reranking model architectures.}
\label{tab:reranker_arch}
\end{table}

\subsection{Loss Function Evolution}

The choice of loss function has a profound impact on reranker behavior, especially at small model scales. Our analysis is consistent with the systematic comparison results from BiXSE and others: at the 500--600M parameter scale, pointwise distillation losses (BCE/MSE) outperform contrastive learning methods in score calibration. Contrastive learning tends to produce clustered scores, making it difficult to set effective thresholds. However, applying contrastive fine-tuning as a second stage (following the DeAR paradigm) can further enhance ranking discrimination without sacrificing calibration quality.

\begin{table}[H]
\centering
\small
\begin{tabularx}{\textwidth}{lXX}
\toprule
\textbf{Loss Type} & \textbf{Advantages} & \textbf{Limitations} \\
\midrule
Pointwise (BCE/MSE) & Stable gradients; 0--1 continuous labels; calibrated prob. & No direct relative ranking opt.; score clustering \\
Pairwise (RankNet~\cite{ranknet}) & Learns relative ranking; effective for hard negatives & $O(n^2)$ complexity; no list-level optimization \\
Listwise (InfoNCE) & Optimizes full ranking; in-batch negative sampling & Temperature-sensitive; requires large batches \\
Distillation (KL/MSE) & Transfers LLM reasoning; 0.6B can reach 7B level & Teacher inference overhead; quality bounded by teacher \\
\bottomrule
\end{tabularx}
\caption{Comparison of loss function families for reranker training.}
\label{tab:loss}
\end{table}

\subsection{Agent Memory Systems}

The rapid development of agent memory architectures provides the application context for MemReranker.
MemOS~\cite{memos} proposes a memory operating system that treats memory as a
first-class computational resource, introducing MemCube as a unified abstraction for managing
parametric, activation, and plaintext memory through a three-layer architecture. Memory3~\cite{memory3}
introduces explicit memory as a third form of memory alongside implicit memory (model parameters) and
working memory (context key-values), demonstrating that a 2.4B model with explicit memory can
outperform larger LLMs and RAG models. Mem0~\cite{mem0} provides a production-ready memory layer
that dynamically extracts and retrieves salient information from ongoing conversations, evaluated
on the LOCOMO benchmark~\cite{locomo}. A-Mem~\cite{amem} proposes an agentic semantic memory
system where memories are structured with contextual tags and dynamically linked. HippoRAG~\cite{hipporag}
draws inspiration from hippocampal memory indexing theory for long-term memory in LLMs.

On the evaluation side, the LOCOMO benchmark provides the primary testbed for
very long-term conversational memory, encompassing question answering, event summarization, and
multi-modal dialogue generation across 300-turn conversations spanning up to 35 sessions.
HaluMem~\cite{halumem} introduces the first operation-level hallucination evaluation benchmark
tailored to memory systems, decomposing memory workflows into extraction, updating, and
question answering stages. MemRL~\cite{memrl} further demonstrates that memory retrieval quality
is critical for agent self-evolution, proposing reinforcement learning on episodic memory to
replace passive semantic matching.

These systems and benchmarks collectively establish that \emph{retrieval quality is the bottleneck
of agent memory} a problem that MemReranker directly addresses through reasoning-capable,
calibrated reranking.

\section{Method}

\subsection{Model Architecture and Evaluation Design}

\begin{figure*}[t]
    \centering
    \smartincludegraphics{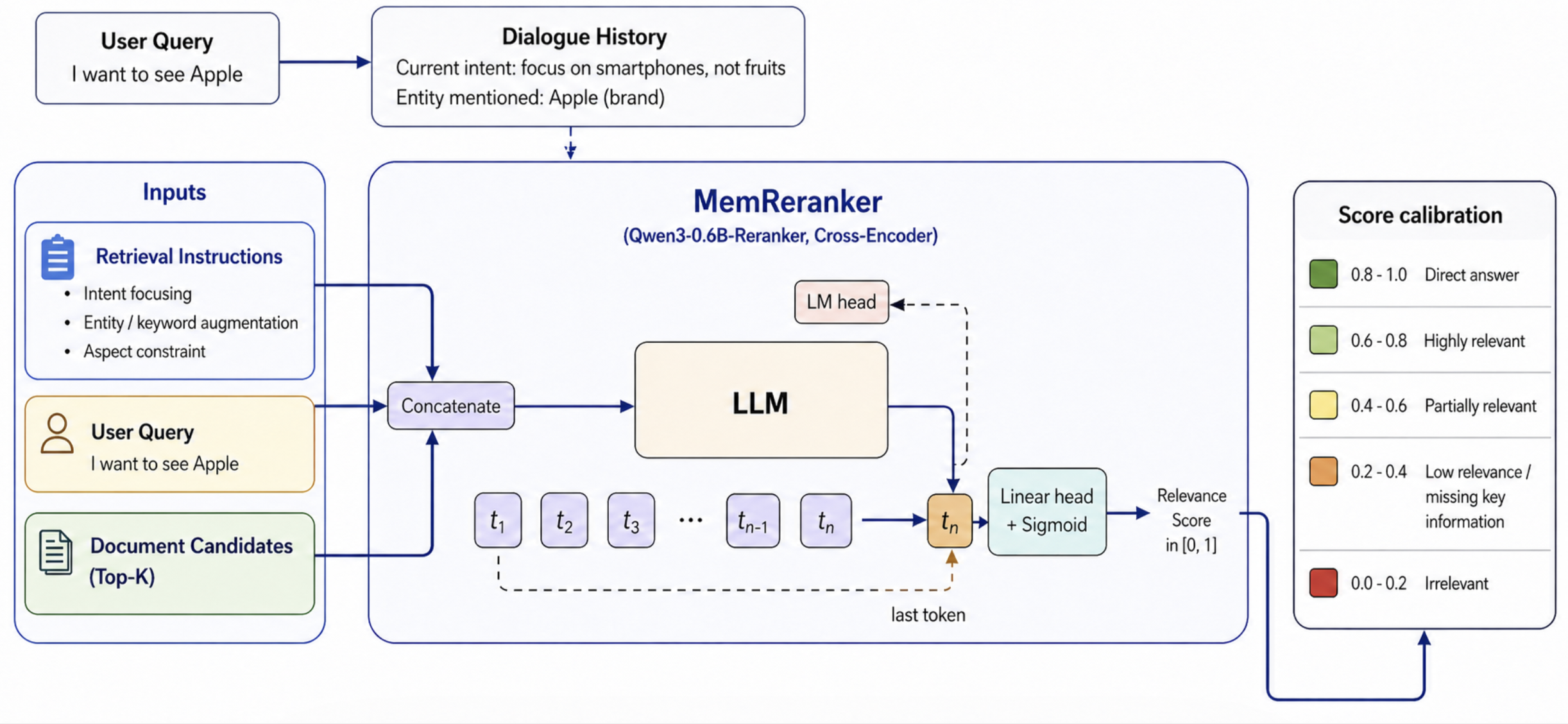}{0.96\textwidth}
    \caption{\textbf{MemReranker architecture.} The model is built on Qwen3-Reranker with BCE loss training. The last-token representation is projected through a linear head and sigmoid activation to produce calibrated $[0,1]$ relevance scores. Three categories of retrieval instructions enable intent focusing, entity augmentation, and aspect-constraint scoring.}
    \label{fig:framework}
\end{figure*}

As illustrated in Figure~\ref{fig:framework}, MemReranker utilizes Qwen3-Reranker as its foundation. We employ Binary Cross-Entropy (BCE) loss for the training process—a design choice informed by the empirical evidence from BiXSE~\cite{bixse}. Their findings demonstrate that at this specific parameter scale, BCE-trained models consistently yield superior performance compared to those trained with InfoNCE loss, effectively establishing BCE as the optimal maximum-likelihood estimator for sigmoid-activated relevance scoring.

\subsubsection{Scoring Mechanism}

The model extracts the last-token representation, passes it through a linear head to obtain a scalar value, and applies sigmoid activation to produce a $[0,1]$ relevance probability. Unlike encoder-based cross-encoders that produce poorly calibrated logits, we establish a hierarchical relevance scoring system. Distillation training produces well-separated scores on a five-level relevance scale: 0--0.2 (irrelevant), 0.2--0.4 (low relevance, missing key information), 0.4--0.6 (partially relevant), 0.6--0.8 (highly relevant), and 0.8--1.0 (direct answer). This calibration directly addresses the left-skewed score distribution problem observed in BGE-Reranker.

\subsubsection{Instruction-Aware Design}

Inspired by the instruction-following capabilities of Qwen3-Reranker and the task-aware approach of Jina Reranker v3~\cite{jinareranker}, MemReranker supports three categories of retrieval instructions:

\paragraph{Intent-Focusing Instructions.}
These extract the core retrieval intent from history-heavy long queries. For example, when dialogue history discusses mobile phone preferences and the current query is ``I want to look at Apple,'' the instruction guides the model to interpret this as a smartphone query rather than a fruit query.

\paragraph{Entity/Keyword Augmentation Instructions.}
These bridge the vocabulary gap between colloquial user queries and professional document terminology, mapping informal descriptions to domain-specific terms.

\paragraph{Aspect-Constraint Instructions.}
When a query contains multiple needs but a document satisfies only one aspect, the instruction guides the model to focus on the relevant portion, enabling partial-match scoring.

\subsubsection{Multi-Dimensional Label Scoring System}

We also design a comprehensive evaluation rubric for post-construction data assessment. Documents are scored through criteria designed to capture aspects that traditional rerankers miss:

\paragraph{Semantic Relevance (30\%).}
Entity consistency (resolving coreferences such as ``it'' to the correct entity) and intent alignment (whether the document responds to the user's operational intent).

\paragraph{Attribute Precision (30\%).}
Feature matching and granularity alignment. A query about ``Beijing'' should not match a document about ``China'' (too broad) or ``Haidian District'' (too narrow).

\paragraph{Information Completeness (25\%).}
Direct answer coverage, reasoning support quality, and partial-answer penalties.

\paragraph{Answer Density and Signal-to-Noise Ratio (15\%).}
Whether key information is prominent or buried in noise, and the document's self-sufficiency.

\subsection{Training Pipeline}

\begin{figure}[t]
    \centering
    \smartincludegraphics{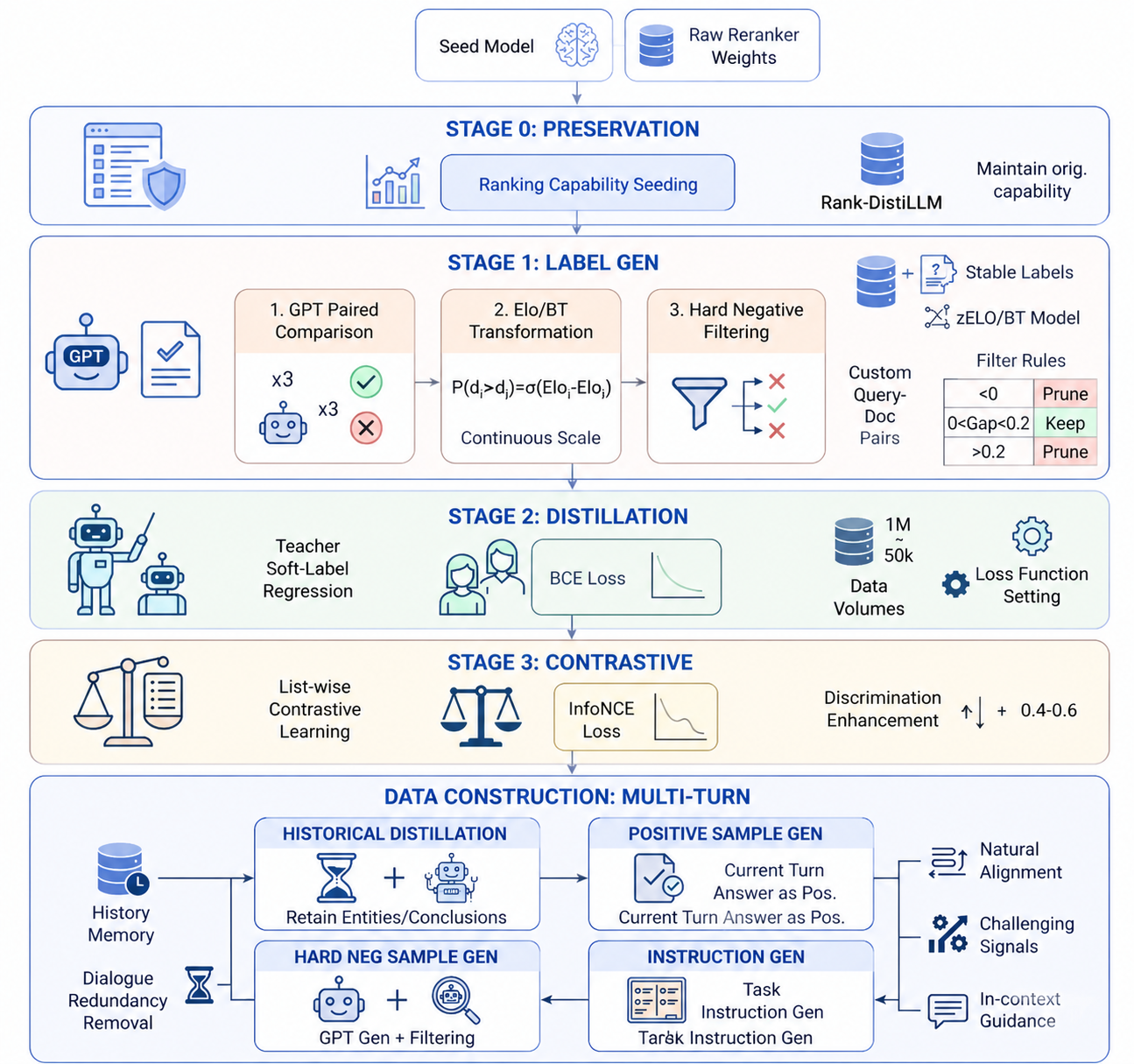}{0.8\linewidth}
    \caption{\textbf{MemReranker training pipeline.} A progressive distillation strategy that transfers LLM reasoning and scoring capabilities to the student model through carefully designed staged learning objectives: Stage~0 (general capability preservation), Stage~1 (teacher label generation), Stage~2 (pointwise BCE distillation), and Stage~3 (contrastive InfoNCE fine-tuning).}
    \label{fig:pipeline}
\end{figure}

Our training pipeline implements a progressive distillation strategy that gradually transfers LLM reasoning and scoring capabilities to the student model through carefully designed staged learning objectives. Table~\ref{tab:pipeline} summarizes the full pipeline.

\begin{table}[H]
\centering
\small
\begin{adjustbox}{max width=\linewidth}
\begin{tabular}{llll}
\toprule
\textbf{Stage} & \textbf{Core Method} & \textbf{Data Source / Scale} & \textbf{Key Details} \\
\midrule
\textbf{Stage 0:} General & Initialize ranking with & Rank-DistiLLM dataset & Preserve original Qwen- \\
capability preservation & public distillation data & & Reranker capabilities \\
\midrule
\textbf{Stage 1:} Teacher & Pairwise LLM comparison: & Custom query-document & Majority voting ($\times 3$) \\
label generation & GPT + Qwen ensemble evaluates doc pairs & pairs + Qwen-Reranker data & ensures label stability \\
 & Elo/BT score conversion & & Bradley-Terry model \\
 & Hard negative filtering & & Cosine gap analysis + \\
 & & & cross-verification \\
\midrule
\textbf{Stage 2:} Pointwise & BCE loss regression on & $\sim$1M general-domain pairs & BiXSE shows BCE $>$ MSE \\
distillation & teacher soft labels & + $\sim$50K multi-turn synthetic & and InfoNCE at this scale \\
\midrule
\textbf{Stage 3:} Contrastive & Listwise InfoNCE loss & [query, pos, neg$_1$, neg$_2$, \ldots] & Enhances discrimination in \\
fine-tuning & (DeAR two-stage paradigm) & tuples & the 0.4--0.6 relevance band \\
\midrule
\textbf{Data:} Multi-turn & History distillation: LLM & Memory-oriented & Remove dialogue redundancy; \\
dialogue construction & condenses prior turns & conversational data & generate hard negatives \\
 & to core entities/conclusions & & with GPT + cross-verification \\
\bottomrule
\end{tabular}
\end{adjustbox}
\caption{Overview of the MemReranker training pipeline, from general capability preservation through multi-turn dialogue data construction.}
\label{tab:pipeline}
\end{table}

\paragraph{Stage 0 --- General Capability Preservation.}
We initialize ranking capabilities using public distillation datasets (Rank-DistiLLM) to preserve the original Qwen-Reranker model's general-purpose abilities before domain adaptation.

\paragraph{Stage 1 --- Teacher Label Generation.}
We use GPT and Qwen ensemble models as teachers, performing pairwise document comparisons with majority voting ($\times 3$) to ensure label stability. Pairwise comparisons are aggregated into continuous Elo scores via Bradley-Terry modeling~\cite{bradleyterry}, then normalized to $[0,1]$. Following the zELO approach, each document is treated as a ``player,'' and the probability of document $d_i$ being preferred over $d_j$ is modeled as:
\begin{equation}
    P(d_i \succ d_j) = \sigma(\text{Elo}_i - \text{Elo}_j),
\end{equation}
where $\sigma(\cdot)$ is the sigmoid function. Maximum likelihood estimation fits absolute Elo scores from sparse pairwise comparisons, converting \emph{relative pairwise preferences into continuous absolute scores}.

For hard negative filtering, we use cosine similarity gap analysis with BGE-Reranker-v2-m3 cross-verification: gap $< 0$: cross-verify and remove annotation errors; $0 <$ gap $< 0.2$: retain as hard negatives; gap $> 0.2$: retain only 20\% as easy negatives.

\paragraph{Stage 2 --- Pointwise Distillation.}
The student model regresses teacher soft labels using BCE loss on approximately 1M general-domain pairs and 50K multi-turn synthetic dialogue pairs. Following BiXSE's findings, BCE loss is preferred over MSE and InfoNCE at this parameter scale.

\paragraph{Stage 3 --- Contrastive Fine-Tuning.}
Listwise contrastive learning (InfoNCE loss) further enhances the model's discrimination in the 0.4--0.6 relevance band, addressing the score clustering issue inherent in pointwise training.

\paragraph{Multi-Turn Dialogue Data Construction.}
For the $n$-th turn query, the preceding $n{-}1$ turns are distilled by an LLM to retain only core entities and conclusions, removing dialogue redundancy. Current-turn answers serve directly as positive-sample documents. Hard negatives are generated via the same multi-teacher ensemble (GPT and Qwen) with cosine similarity gap analysis and BGE-Reranker cross-verification. Task instructions are generated based on query-document pair characteristics for inference-time context guidance.

Synthesizing the above reference works, we distill three key design principles: (1) at small parameter scales ($<$1B), BCE pointwise distillation outperforms contrastive learning (from BiXSE's findings); (2) two-stage training (pointwise first, then listwise) balances calibration quality with ranking discrimination (from DeAR's paradigm); and (3) Elo/BT-based score conversion provides a principled bridge from pairwise comparisons to continuous scores . MemReranker's training pipeline is a unified application of these three principles.

\section{Experiments}

We evaluate MemReranker across several dimensions:

\begin{itemize}[nosep]
    \item \textbf{Memory scenario benchmark:} The LOCOMO/LongMemEval dataset for testing memory dialogue retrieval capabilities.
    \item \textbf{Hard-case evaluation:} Constructed hard-to-distinguish data types reflecting model discrimination on complex samples (built by Opus-4.6).
    \item \textbf{Finance/Healthcare benchmarks:} Vertical-domain evaluation to ensure the model retains basic ranking ability and to compare against mainstream models.
\end{itemize}

\subsection{Experimental Setup}

\paragraph{Baselines.}
We compare MemReranker against two categories of baselines.
\textit{Dedicated rerankers} include BGE-Reranker-v2-m3 as the primary baseline, along with Qwen3-Reranker-4B and Qwen3-Reranker-8B representing state-of-the-art open-source rerankers at larger parameter scales.
\textit{LLM-as-judge rerankers} include GPT-4o-mini and Gemini-3-Flash, which perform pointwise relevance scoring via prompting and serve as upper-bound references for reasoning-intensive ranking.

\paragraph{Evaluation Metrics.}
We adopt standard information retrieval metrics throughout our experiments. \textbf{Mean Average Precision (MAP)} computes the mean of average precision across all queries:
\begin{equation}
    \mathrm{MAP} = \frac{1}{|Q|}\sum_{q=1}^{|Q|}\frac{1}{|\mathrm{Rel}_q|}\sum_{k=1}^{n} P_q(k)\cdot \mathbb{1}[r_q(k)\in \mathrm{Rel}_q],
\end{equation}
where $P_q(k)$ is the precision at rank $k$ for query $q$, $\mathrm{Rel}_q$ is the set of relevant documents, and $\mathbb{1}[\cdot]$ is the indicator function.
\textbf{Mean Reciprocal Rank (MRR)} measures the reciprocal of the rank of the first relevant document:
\begin{equation}
    \mathrm{MRR} = \frac{1}{|Q|}\sum_{q=1}^{|Q|}\frac{1}{\mathrm{rank}_q},
\end{equation}
where $\mathrm{rank}_q$ is the position of the first relevant document for query $q$.
\textbf{Normalized Discounted Cumulative Gain (NDCG@$k$)} evaluates graded relevance with position-based discounting:
\begin{equation}
    \mathrm{NDCG@}k = \frac{\mathrm{DCG@}k}{\mathrm{IDCG@}k}, \quad \mathrm{DCG@}k = \sum_{i=1}^{k}\frac{2^{\mathrm{rel}_i}-1}{\log_2(i+1)},
\end{equation}
where $\mathrm{rel}_i$ is the relevance grade of the document at rank $i$ and $\mathrm{IDCG@}k$ is the ideal DCG obtained by sorting all documents by decreasing relevance.
We report NDCG at cutoffs $k \in \{1, 3, 10\}$ as well as the full-ranking NDCG.
Additionally, we report \textbf{Recall@$k$} ($k \in \{3, 5, 20\}$) measuring the fraction of relevant documents retrieved within the top-$k$ results, and \textbf{F1} as the harmonic mean of precision and recall.

\paragraph{Training Configuration.}
Both models are initialized from corresponding Qwen3-Reranker checkpoints and trained on 8$\times$A800 (80\,GB) GPUs for 3 epochs with AdamW (lr $2\times10^{-5}$) and gradient checkpointing.
MemReranker-0.6B uses ZeRO-Stage~0, per-device batch size 16, gradient accumulation 4 (effective batch 512), and max length 8192.
MemReranker-4B uses ZeRO-Stage~2, per-device batch size 4, gradient accumulation 8 (effective batch 256), and max length 8192. Although the original Qwen3 architecture supports context lengths up to 32K tokens, we limit the maximum sequence length to 8,192 due to training resource constraints and the length distribution of our training data. The best checkpoint is selected by validation loss evaluated every 0.5 epochs.

\subsection{Memory Scenario Evaluation}
\subsubsection{LOCOMO}

To avoid interference from intermediate pipeline components, we directly use concatenated dialogues as raw memories and employ a recall + rerank approach, uniformly using BGE-M3 Top-100 as candidate results to ensure consistent input across all reranking models.

\begin{table}[H]
\centering
\small
\begin{adjustbox}{max width=\linewidth}
\begin{tabular}{l cccccccccc}
\toprule
\textbf{Model} & \textbf{MAP} & \textbf{MRR} & \textbf{NDCG@1} & \textbf{NDCG@3} & \textbf{NDCG@10} & \textbf{NDCG} & \textbf{R@3} & \textbf{R@5} & \textbf{R@20} & \textbf{F1} \\
\midrule
BGE-v2-m3        & 0.6708 & 0.6994 & 0.6071 & 0.6718 & 0.7140 & 0.7358 & 0.7156 & 0.7683 & 0.8631 & 0.5038 \\
Qwen3-Reranker-0.6B & 0.6427 & 0.6727 & 0.5760 & 0.6382 & 0.6889 & 0.7144 & 0.6811 & 0.7479 & 0.8568 & 0.4720 \\
Qwen3-Reranker-4B       & 0.6894 & 0.7162 & 0.6234 & 0.6906 & 0.7324 & 0.7504 & 0.7353 & 0.7958 & 0.8726 & 0.5221 \\
Qwen3-Reranker-8B       & 0.7206 & 0.7484 & 0.6656 & 0.7240 & 0.7593 & 0.7748 & 0.7633 & 0.8130 & 0.8800 & 0.5518 \\
GPT-4o-mini       & 0.7151 & 0.7415 & 0.6565 & 0.7186 & 0.7526 & 0.7698 & 0.7602 & 0.8122 & 0.8678 & 0.5441 \\
Gemini-3-Flash    & \textbf{0.7772} & \textbf{0.7973} & \textbf{0.7370} & \textbf{0.7780} & \textbf{0.8068} & \textbf{0.8161} & \textbf{0.8047} & \textbf{0.8469} & \textbf{0.8919} & \textbf{0.6218} \\
\midrule
\rowcolor{blue!8}
MemReranker-0.6B    & \underline{0.7150} & \underline{0.7377} & 0.6500 & 0.7166 & 0.7540 & 0.7696 & 0.7575 & 0.8090 & \underline{0.8848} & 0.5548 \\
\rowcolor{blue!8}
MemReranker-4B      & \underline{0.7366} & \underline{0.7596} & \underline{0.6786} & \underline{0.7391} & \underline{0.7734} & \underline{0.7861} & \underline{0.7774} & \underline{0.8239} & \underline{0.8886} & \underline{0.5768} \\
\bottomrule
\end{tabular}
\end{adjustbox}
\caption{LOCOMO memory retrieval benchmark results. Best values are \textbf{bold}; second-best are \underline{underlined}. \colorbox{blue!8}{Shaded rows} indicate our models. MemReranker-4B achieves the highest scores among all dedicated rerankers, while MemReranker-0.6B with only 0.6B parameters already matches GPT-4o-mini. Detailed per-category results are provided in Appendix~\ref{app:locomo_cat}.}
\label{tab:locomo}
\end{table}

As shown in Table~\ref{tab:locomo}, MemReranker-0.6B with only 0.6B parameters substantially outperforms BGE-Reranker-v2-m3 (MAP 0.7150 vs.\ 0.6708, a 4.4- percentage point improvement). More critically, it also surpasses the much larger Qwen3-Reranker-4B (0.6894), demonstrating that MemReranker's architecture is effectively optimized for memory retrieval scenarios. Among non-LLM-as-judge dedicated rerankers, MemReranker-4B achieves the highest scores across all metrics; Qwen3-Reranker-8B, with twice the parameters, only reaches MAP 0.7206, lagging by 1.6 points, further demonstrating MemReranker's parameter efficiency advantage. GPT-4o-mini's MAP is essentially on par with MemReranker-0.6B, while MemReranker-4B clearly leads. Considering GPT-4o-mini's substantially higher inference cost and latency, MemReranker offers a significant cost-effectiveness advantage. Gemini-3-Flash achieves the overall best results, but as a larger model with prohibitively high deployment costs, MemReranker as a lightweight dedicated model matches it on several metrics while offering substantial latency and cost advantages in practical deployment.

\subsubsection{LongMemEval}
 
To further validate MemReranker's effectiveness across diverse long-term memory retrieval patterns, we evaluate on the LongMemEval benchmark, which comprises 500 queries spanning six categories: knowledge update, multi-session reasoning, single-session assistant/user/preference recall, and temporal reasoning. For LongMemEval datasets we use BGE-M3 Top-50 as the unified candidate set for all reranking models.
 
\begin{table}[H]
\centering
\small
\begin{adjustbox}{max width=\linewidth}
\begin{tabular}{l cccccccccc}
\toprule
\textbf{Model} & \textbf{MAP} & \textbf{MRR} & \textbf{NDCG@1} & \textbf{NDCG@3} & \textbf{NDCG@10} & \textbf{NDCG} & \textbf{R@3} & \textbf{R@5} & \textbf{R@20} & \textbf{F1} \\
\midrule
BGE-v2-m3        & 0.7069 & 0.7510 & 0.6600 & 0.7051 & 0.7573 & 0.7696 & 0.7412 & 0.8095 & 0.8874 & 0.5634 \\
Qwen3-Reranker-0.6B & 0.6955 & 0.7367 & 0.6260 & 0.6900 & 0.7520 & 0.7616 & 0.7313 & 0.8135 & 0.8900 & 0.5567 \\
Qwen3-Reranker-4B       & 0.6501 & 0.6937 & 0.5640 & 0.6403 & 0.7185 & 0.7290 & 0.6872 & 0.7900 & 0.8934 & 0.4981 \\
Qwen3-Reranker-8B       & 0.6966 & 0.7340 & 0.6300 & 0.6851 & 0.7551 & 0.7617 & 0.7210 & 0.8108 & 0.8931 & 0.5490 \\
GPT-4o-mini\tablefootnote{GPT-4o-mini evaluated 481 out of 500 queries on LongMemEval (19 skipped due to API failures).}
                  & 0.5684 & 0.6385 & 0.5322 & 0.5551 & 0.6265 & 0.6676 & 0.5781 & 0.6463 & 0.8349 & 0.4444 \\
Gemini-3-Flash    & 0.7259 & 0.7575 & 0.6660 & 0.7222 & 0.7733 & 0.7808 & 0.7556 & 0.8280 & 0.8977 & 0.6010 \\
\midrule
\rowcolor{blue!8}
MemReranker-0.6B  & \underline{0.7538} & \underline{0.7876} & \underline{0.7020} & \underline{0.7562} & \underline{0.7987} & \underline{0.8023} & \underline{0.7872} & \underline{0.8508} & \underline{0.8969} & \underline{0.6375} \\
\rowcolor{blue!8}
MemReranker-4B    & \textbf{0.8043} & \textbf{0.8315} & \textbf{0.7800} & \textbf{0.8069} & \textbf{0.8354} & \textbf{0.8369} & \textbf{0.8203} & \textbf{0.8734} & \textbf{0.8960} & \textbf{0.7039} \\
\bottomrule
\end{tabular}
\end{adjustbox}
\caption{LongMemEval memory retrieval benchmark results (n=500). Best values are \textbf{bold}; second-best are \underline{underlined}. \colorbox{blue!8}{Shaded rows} indicate our models. MemReranker-4B achieves the best results across all metrics, surpassing both dedicated rerankers and LLM-as-judge baselines including Gemini-3-Flash.}
\label{tab:longmemeval}
\end{table}
 
As shown in Table~\ref{tab:longmemeval}, the LongMemEval benchmark reveals even more pronounced advantages for MemReranker compared to the LOCOMO results. MemReranker-4B achieves the highest scores across all metrics, with MAP 0.8043 surpassing Gemini-3-Flash (0.7259) by 7.8 percentage points, a notably larger margin than observed on LOCOMO. MemReranker-0.6B likewise outperforms all baselines including Gemini-3-Flash, achieving MAP 0.7538 versus 0.7259, demonstrating that even the compact 0.6B model has internalized sufficient reasoning capability for complex memory retrieval. GPT-4o-mini performs substantially worse on this benchmark (MAP 0.5684), suggesting that its prompting-based scoring struggles with the more diverse query patterns in LongMemEval. Qwen3-Reranker-4B and Qwen3-Reranker-8B also lag behind their LOCOMO performance, further highlighting that MemReranker's memory-specific training pipeline generalizes effectively to different memory evaluation settings.

\subsection{Hard-Case Evaluation}

Standard test samples mostly come from public data, risking inflated metrics due to test-set leakage. To test whether the reranking model can effectively transfer the original large model's discrimination ability on hard samples, we use Opus-4.6 to generate a batch of semantically hard-to-distinguish ranking data covering multi-hop reasoning, numerical reasoning, temporal interference, and other complex types.

\begin{table}[H]
\centering
\small
\begin{adjustbox}{max width=\linewidth}
\begin{tabular}{l ccccccccc}
\toprule
\textbf{Model} & \textbf{MAP} & \textbf{MRR} & \textbf{NDCG@1} & \textbf{NDCG@3} & \textbf{NDCG} & \textbf{R@3} & \textbf{R@5} & \textbf{F1@3} & \textbf{F1@5} \\
\midrule
BGE-v2-m3        & 0.797 & 0.915 & 0.850 & 0.824 & 0.929 & 0.456 & 0.700 & 0.570 & 0.700 \\
Qwen3-Reranker-0.6B     & 0.794 & 0.926 & 0.869 & 0.830 & 0.934 & 0.452 & 0.690 & 0.565 & 0.690 \\
Qwen3-Reranker-8B       & 0.840 & 0.955 & 0.927 & 0.872 & 0.952 & 0.484 & 0.746 & 0.605 & 0.746 \\
GPT-4o-mini       & \textbf{0.910} & 0.978 & 0.951 & \textbf{0.933} & \textbf{0.973} & \textbf{0.542} & \textbf{0.828} & \textbf{0.677} & \textbf{0.828} \\
Gemini-3-Flash    & \textbf{0.912} & 0.978 & 0.952 & \textbf{0.938} & \textbf{0.974} & \textbf{0.548} & 0.822 & \textbf{0.685} & 0.822 \\
\midrule
\rowcolor{blue!8}
MemReranker-0.6B    & \underline{0.862} & 0.970 & 0.936 & \underline{0.896} & \underline{0.960} & \underline{0.504} & \underline{0.766} & \underline{0.630} & \underline{0.766} \\
\rowcolor{blue!8}
MemReranker-4B      & 0.893 & \textbf{0.985} & \textbf{0.955} & 0.921 & \underline{0.968} & \underline{0.540} & \underline{0.788} & \underline{0.675} & \underline{0.788} \\
\bottomrule
\end{tabular}
\end{adjustbox}
\caption{Hard-case evaluation results (n=100 per model). Test samples are generated by Opus-4.6 covering multi-hop reasoning, numerical reasoning, temporal interference, and other challenging types. Best values are \textbf{bold}; second-best are \underline{underlined}. \colorbox{blue!8}{Shaded rows} indicate our models.}
\label{tab:hardcase}
\end{table}

As shown in Table~\ref{tab:hardcase}, Gemini-3-Flash and GPT-4o-mini as large models still hold advantages across most metrics. However, after pipeline training, MemReranker's overall metrics are substantially higher than both BGE-Reranker and the original Qwen3-Reranker. We also compared against the original Qwen3-Reranker-8B, which shows no clear advantage on hard samples. MemReranker matches GPT-4o-mini and the stronger Gemini-3-Flash model on key metrics such as NDCG, demonstrating that pipeline training yields significant improvements on complex ranking samples. A detailed per-category analysis is provided in Appendix~\ref{app:hardcase}.

\subsubsection{Case Studies}

\paragraph{Case 1 --- High Lexical Overlap, Low Semantic Relevance.}

Given the query ``How is the battery safety of new energy vehicles?'' and candidate documents, MemReranker-4B (NDCG = 0.991) and MemReranker-0.6B (NDCG = 0.986) correctly identify doc~0 (thermal runaway temperature data) and doc~4 (spontaneous combustion rate statistics), hitting the core safety evidence chain in top-3. GPT-4o-mini (NDCG = 0.926) incorrectly ranks doc~1 (about CATL's new battery energy density improvements) first, misled by the phrase ``safety testing'' when the document's main topic is energy density. BGE-Reranker (NDCG = 0.894) and Qwen3-Reranker-4B (NDCG = 0.886) both incorrectly rank doc~7 (circuit disconnection protection in collisions) first, missing the most direct safety-related thermal runaway data.

\paragraph{Case 2 --- Low Lexical Overlap, High Semantic Relevance.}

Given the query ``How to make a team more cohesive?'' and a candidate document about organizing non-work team activities (outdoor excursions, dinners, game nights) alongside establishing transparent information-sharing mechanisms and fair benefit-distribution systems, MemReranker-0.6B correctly ranks this document first (NDCG = 0.993), while GPT-4o-mini incorrectly places a psychology research document at rank 1 (NDCG = 0.882), and BGE-Reranker ranks an irrelevant corporate culture case study first (NDCG = 0.812).

\paragraph{Case 3 --- Multi-Hop Reasoning.}

Given the query ``What nationality is the author of \emph{One Hundred Years of Solitude}?'', the correct answer requires two-hop reasoning: query $\to$ ``author of \emph{One Hundred Years of Solitude}'' $\to$ doc~2 (reveals Garc\'ia M\'arquez) $\to$ ``What nationality is M\'arquez?'' $\to$ doc~1 (reveals Colombian). MemReranker's top-3 results are $[2, 1, 8]$, correctly identifying both key nodes of the information chain plus supplementary doc~8 (M\'arquez's passing in Mexico City, with Colombia mourning nationally). Other models achieve only a 1/3 top-3 hit rate. Models without internal knowledge or reasoning capability cannot associate the internal knowledge chain required for this query.

\subsection{Vertical-Domain Evaluation}

To ensure the model retains generalization capabilities on par with mainstream ranking models, we also evaluate on finance and healthcare domain benchmark datasets.

\paragraph{Finance.}
We use the FinanceMTEB/FinFact-reranking dataset as the evaluation benchmark.

\begin{table}[H]
\centering
\small
\begin{adjustbox}{max width=\linewidth}
\begin{tabular}{l ccccccc}
\toprule
\textbf{Model} & \textbf{MAP} & \textbf{MRR} & \textbf{NDCG@1} & \textbf{NDCG@3} & \textbf{NDCG@5} & \textbf{Recall} & \textbf{F1} \\
\midrule
BGE-v2-m3        & 0.991 & 0.991 & 0.985 & 0.991 & 0.993 & 0.963 & 0.971 \\
Qwen3-Reranker-8B       & \textbf{0.997} & \textbf{0.997} & \textbf{0.995} & 0.997 & 0.997 & \textbf{0.995} & \textbf{0.995} \\
GPT-4o-mini       & 0.995 & 0.995 & 0.991 & 0.997 & 0.997 & 0.942 & 0.954 \\
\midrule
\rowcolor{blue!8}
MemReranker-0.6B    & \textbf{0.997} & \textbf{0.997} & \textbf{0.995} & \textbf{0.998} & \textbf{0.998} & 0.968 & 0.977 \\
\rowcolor{blue!8}
MemReranker-4B      & 0.993 & 0.993 & 0.990 & 0.993 & 0.995 & 0.990 & 0.990 \\
\bottomrule
\end{tabular}
\end{adjustbox}
\caption{FinFact reranking benchmark results (finance domain, n=1000). Best values are \textbf{bold}. \colorbox{blue!8}{Shaded rows} indicate our models.}
\label{tab:finfact}
\end{table}

\paragraph{Healthcare.}
We evaluate on three medical-domain datasets: \textbf{NFCorpus} (English, $\sim$323 test queries, 3600 documents, nutrition and medicine), \textbf{SciFact} (English, 300 test queries, 5200 documents, biomedical claim verification), and \textbf{CMedQAv2} (Chinese, 4000 test samples, community medical QA).

\begin{table}[H]
\centering
\small
\setlength{\tabcolsep}{3.5pt}
\begin{adjustbox}{max width=\linewidth}
\begin{tabular}{ll cccccc}
\toprule
\textbf{Dataset} & \textbf{Model} & \textbf{MAP} & \textbf{MRR} & \textbf{NDCG@1} & \textbf{NDCG@5} & \textbf{NDCG@10} & \textbf{R@10} \\
\midrule
\multirow{5}{*}{\textbf{NFCorpus}}
& BGE-Reranker-v2-m3  & 0.6510 & 0.8482 & 0.7926 & 0.7151 & 0.7084 & 0.4026 \\
& Qwen3-Reranker-8B    & 0.7363 & 0.8870 & 0.8424 & \textbf{0.8018} & \textbf{0.7997} & \textbf{0.4651} \\
& GPT-4o-mini           & 0.6141 & 0.8672 & 0.8259 & 0.7217 & 0.6722 & 0.3593 \\
\cmidrule(lr){2-8}
& \cellcolor{blue!8} MemReranker-0.6B & \cellcolor{blue!8} \underline{0.7172} & \cellcolor{blue!8} \underline{0.8873} & \cellcolor{blue!8} \underline{0.8452} & \cellcolor{blue!8} \underline{0.7896} & \cellcolor{blue!8} \underline{0.7738} & \cellcolor{blue!8} \underline{0.4376} \\
& \cellcolor{blue!8} MemReranker-4B   & \cellcolor{blue!8} \textbf{0.7456} & \cellcolor{blue!8} \textbf{0.8927} & \cellcolor{blue!8} \textbf{0.8514} & \cellcolor{blue!8} 0.8082 & \cellcolor{blue!8} 0.8012 & \cellcolor{blue!8} 0.4619 \\
\midrule
\multirow{5}{*}{\textbf{SciFact}}
& BGE-Reranker-v2-m3  & 0.9708 & 0.9713 & 0.9567 & 0.9750 & 0.9762 & 0.9933 \\
& Qwen3-Reranker-8B    & 0.9864 & 0.9872 & 0.9767 & \textbf{0.9901} & 0.9901 & \textbf{1.0000} \\
& GPT-4o-mini           & 0.9234 & 0.9356 & 0.9200 & 0.9266 & 0.9266 & 0.9338 \\
\cmidrule(lr){2-8}
& \cellcolor{blue!8} MemReranker-0.6B & \cellcolor{blue!8} \textbf{0.9871} & \cellcolor{blue!8} \textbf{0.9880} & \cellcolor{blue!8} \textbf{0.9800} & \cellcolor{blue!8} 0.9894 & \cellcolor{blue!8} \textbf{0.9905} & \cellcolor{blue!8} \textbf{1.0000} \\
& \cellcolor{blue!8} MemReranker-4B   & \cellcolor{blue!8} 0.9811 & \cellcolor{blue!8} 0.9811 & \cellcolor{blue!8} 0.9633 & \cellcolor{blue!8} 0.9861 & \cellcolor{blue!8} 0.9861 & \cellcolor{blue!8} \textbf{1.0000} \\
\midrule
\multirow{5}{*}{\textbf{CMedQAv2}}
& BGE-Reranker-v2-m3  & \underline{0.8347} & \underline{0.8615} & \underline{0.7870} & \underline{0.8578} & \underline{0.8745} & \underline{0.9666} \\
& Qwen3-Reranker-8B    & \textbf{0.8536} & \textbf{0.8761} & \textbf{0.8100} & \textbf{0.8742} & \textbf{0.8890} & \textbf{0.9701} \\
& GPT-4o-mini           & 0.6750 & 0.7384 & 0.6430 & 0.7125 & 0.7235 & 0.8101 \\
\cmidrule(lr){2-8}
& \cellcolor{blue!8} MemReranker-0.6B & \cellcolor{blue!8} 0.7590 & \cellcolor{blue!8} 0.7938 & \cellcolor{blue!8} 0.6980 & \cellcolor{blue!8} 0.7864 & \cellcolor{blue!8} 0.8134 & \cellcolor{blue!8} 0.9437 \\
& \cellcolor{blue!8} MemReranker-4B   & \cellcolor{blue!8} 0.7700 & \cellcolor{blue!8} 0.8000 & \cellcolor{blue!8} 0.7240 & \cellcolor{blue!8} 0.7950 & \cellcolor{blue!8} 0.8190 & \cellcolor{blue!8} 0.9350 \\
\bottomrule
\end{tabular}
\end{adjustbox}
\caption{Medical domain benchmark results across three datasets. Best values are \textbf{bold}; second-best are \underline{underlined}. \colorbox{blue!8}{Shaded cells} indicate our models.}
\label{tab:medical}
\end{table}

From the medical-domain results, Qwen3-Reranker-8B holds a clear advantage as a large-parameter ranking model. BGE-Reranker leads on CMedQAv2, but its large-scale training set likely includes the original dataset. On the other two datasets, MemReranker-0.6B/4B already matches or exceeds larger parameter models on several metrics and surpasses closed-source models like GPT-4o-mini.

\subsection{Latency Analysis}

To provide a more intuitive comparison of model latency performance, we record response-time statistics on the test dataset 1k tokens. BGE-Reranker benefits from its encoder architecture for fast response times, while GPT-4o-mini incurs significant latency due to prompt processing and score generation (detailed prompt in Appendix~\ref{app:prompt}). MemReranker-0.6B, leveraging its small parameter count and inheriting the Qwen-Reranker scoring paradigm, achieves approximately 200ms latency, offering an excellent balance of low latency and high accuracy compared to both BGE and large models.

\begin{table}[H]
\centering
\small
\begin{tabular}{lccc}
\toprule
\textbf{Latency Metric} & \textbf{BGE-Reranker-v2-m3} & \textbf{MemReranker-0.6B} & \textbf{GPT-4o-mini} \\
\midrule
Avg (ms) & 241.0 & 247.2 & 1549.2 \\
P50 (ms) & 239.4 & 246.8 & 1500.1 \\
P95 (ms) & 271.4 & 327.4 & 2034.5 \\
P99 (ms) & 271.4 & 327.4 & 2524.2 \\
Min (ms) & 213.1 & 204.5 & 1083.0 \\
\bottomrule
\end{tabular}
\caption{Inference latency comparison across reranking models. MemReranker-0.6B achieves $\sim$3$\times$ the latency of BGE but $\sim$8$\times$ faster than GPT-4o-mini, with substantially better ranking quality than BGE.}
\label{tab:latency}
\end{table}

\FloatBarrier

\section{Conclusion}

This paper introduces the MemReranker family of reranking models and reframes agent memory retrieval from generic semantic matching to reasoning-capable, instruction-aware, and calibrated reranking. MemReranker-0.6B demonstrates that a multi-stage distillation pipeline can transfer LLM-level reasoning into a compact model, achieving ranking quality on par with GPT-4o-mini at 8$\times$ lower latency. MemReranker-4B further approaches or exceeds Gemini-3-Flash on hard-case scenarios while maintaining practical deployment costs.

Experiments on the LOCOMO memory retrieval benchmark, hard-case benchmarks, and vertical-domain datasets (finance, healthcare) show clear gains, particularly in scenarios requiring multi-hop reasoning, temporal understanding, and causal logic---precisely the areas where conventional cross-encoder rerankers fail. The calibrated $[0,1]$ scoring with well-separated relevance levels directly addresses the practical challenge of threshold-based filtering in production memory systems.

Several directions remain open: extending instruction-aware capabilities to support more complex multi-turn dialogue patterns; jointly optimizing the recall and reranking stages; evaluating stability in online deployment with real user traffic; and exploring larger teacher ensembles for further distillation quality improvements. We hope this work provides a practical path toward deploying reasoning-capable rerankers in production agent memory systems.

\bibliographystyle{unsrt}
\bibliography{main}

\appendix

\section{Hard-Case Category Analysis}
\label{app:hardcase}

Table~\ref{tab:ndcg_detail} presents a fine-grained NDCG breakdown by difficulty category.

\begin{table}[H]
\centering
\small
\begin{adjustbox}{max width=\linewidth}
\begin{tabular}{lcccc}
\toprule
\textbf{Category} & \textbf{BGE-v2-m3} & \textbf{MemReranker-0.6B} & \textbf{Qwen3-R-0.6B} & \textbf{GPT-4o-mini} \\
\midrule
Low lexical overlap, high semantic   & 0.897 & \textbf{0.976} & 0.927 & 0.961 \\
Paraphrase interference              & 0.978 & 0.992 & 0.978 & \textbf{0.993} \\
Negation / semantic reversal          & 0.950 & 0.964 & 0.947 & \textbf{0.988} \\
Causal reversal                       & 0.950 & \textbf{0.975} & 0.948 & 0.974 \\
Multi-hop reasoning                   & 0.904 & 0.923 & 0.910 & \textbf{0.988} \\
Entity confusion                      & 0.944 & \textbf{0.973} & 0.965 & 0.970 \\
Coreference resolution                & 0.874 & 0.934 & 0.866 & \textbf{0.952} \\
Numerical reasoning                   & 0.869 & 0.942 & 0.907 & \textbf{0.961} \\
Temporal interference                 & 0.942 & 0.946 & 0.935 & \textbf{0.975} \\
Conditional constraint interference   & 0.957 & 0.971 & 0.963 & \textbf{0.985} \\
Granularity mismatch                  & 0.967 & 0.980 & 0.969 & \textbf{0.990} \\
Opinion vs.\ fact                     & 0.903 & \textbf{0.930} & 0.878 & 0.929 \\
Implicit intent                       & 0.960 & 0.975 & 0.968 & \textbf{0.982} \\
High lexical overlap, low semantic    & 0.932 & 0.975 & 0.926 & \textbf{0.979} \\
\bottomrule
\end{tabular}
\end{adjustbox}
\caption{Per-category NDCG breakdown on hard-case evaluation. Overall ranking: GPT-4o-mini $\approx$ MemReranker-0.6B $>$ BGE-Reranker-v2-m3 $\approx$ Qwen3-Reranker-0.6B.}
\label{tab:ndcg_detail}
\end{table}

GPT-4o-mini performs best in most categories, particularly in negation/semantic reversal (0.988), multi-hop reasoning (0.988), and granularity mismatch (0.990), reflecting the ceiling capability of large models in deep semantic understanding.

MemReranker-0.6B closely follows, even surpassing GPT-4o-mini in low-lexical-overlap-high-semantic (0.976 vs.\ 0.961) and high-lexical-overlap-low-semantic (0.975 vs.\ 0.979) scenarios, demonstrating excellent lexical-semantic decoupling. On paraphrase interference (0.992 vs.\ 0.993), causal reversal (0.975 vs.\ 0.974), and entity confusion (0.973 vs.\ 0.970), the gap is within 0.003.

\textbf{MemReranker's ranking capability is essentially on par with GPT-4o-mini.} Although GPT-4o-mini still leads by approximately 3--6\% in scenarios requiring complex reasoning (multi-hop reasoning, negation semantics), MemReranker---as a 0.6B-parameter lightweight model with less than one percent of GPT-4o-mini's parameters---achieves comparable ranking quality and even outperforms on certain surface-level semantic interference scenarios. Considering inference cost and latency, MemReranker offers a significant cost-effectiveness advantage in practical deployments.

\section{Ranking Prompt Template}
\label{app:prompt}

\begin{promptbox}{LLM Ranking Prompt (used for teacher label generation)}
\begin{lstlisting}[style=promptlisting]
You are a document relevance ranking expert. Given a query
and a set of documents, rank the document IDs by relevance
to the query from highest to lowest.

Query: {query}

Document list:
{doc_list}

Output the ranked document IDs as a JSON array,
e.g. [3, 0, 2, 1, ...].
Output only the JSON array, nothing else.
\end{lstlisting}
\end{promptbox}

\section{LOCOMO Category Results}
\label{app:locomo_cat}

\subsection{Category 1: Single-Hop (n=282)}

\begin{table}[H]
\centering
\small
\begin{adjustbox}{max width=\linewidth}
\begin{tabular}{lccccccc}
\toprule
\textbf{LOCOMO} & \textbf{BGE-v2-m3} & \textbf{MemReranker-0.6B} & \textbf{MemReranker-4B} & \textbf{Qwen3-R-4B} & \textbf{Qwen3-R-8B} & \textbf{GPT-4o-mini} & \textbf{Gemini-3-Flash} \\
\midrule
MAP & 0.5387 & 0.5848 & 0.6337 & 0.5749 & 0.5959 & 0.5431 & \textbf{0.6630} \\
MRR & 0.6521 & 0.6754 & 0.7336 & 0.6892 & 0.7048 & 0.6615 & \textbf{0.7512} \\
Recall@5 & 0.6025 & 0.6650 & 0.6945 & 0.6438 & 0.6700 & 0.6300 & \textbf{0.7389} \\
Recall@10 & 0.7166 & 0.7669 & 0.7818 & 0.7477 & 0.7638 & 0.7310 & \textbf{0.8289} \\
\bottomrule
\end{tabular}
\end{adjustbox}
\caption{LOCOMO Category 1: Single-hop queries (n=282).}
\end{table}

\subsection{Category 2: Temporal (n=321)}

\begin{table}[H]
\centering
\small
\begin{adjustbox}{max width=\linewidth}
\begin{tabular}{lccccccc}
\toprule
\textbf{LOCOMO} & \textbf{BGE-v2-m3} & \textbf{MemReranker-0.6B} & \textbf{MemReranker-4B} & \textbf{Qwen3-R-4B} & \textbf{Qwen3-R-8B} & \textbf{GPT-4o-mini} & \textbf{Gemini-3-Flash} \\
\midrule
MAP & 0.7489 & 0.7811 & 0.8121 & 0.7464 & 0.7737 & 0.7983 & \textbf{0.8472} \\
MRR & 0.7621 & 0.7899 & 0.8204 & 0.7578 & 0.7913 & 0.8026 & \textbf{0.8549} \\
Recall@5 & 0.8229 & 0.8692 & 0.8853 & 0.8437 & 0.8588 & 0.8712 & \textbf{0.8925} \\
Recall@10 & 0.8629 & 0.8915 & \textbf{0.8962} & 0.8707 & 0.8780 & 0.8837 & 0.8962 \\
\bottomrule
\end{tabular}
\end{adjustbox}
\caption{LOCOMO Category 2: Temporal queries (n=321).}
\end{table}

\subsection{Category 3: Multi-Hop (n=96)}

\begin{table}[H]
\centering
\small
\begin{adjustbox}{max width=\linewidth}
\begin{tabular}{lccccccc}
\toprule
\textbf{LOCOMO} & \textbf{BGE-v2-m3} & \textbf{MemReranker-0.6B} & \textbf{MemReranker-4B} & \textbf{Qwen3-R-4B} & \textbf{Qwen3-R-8B} & \textbf{GPT-4o-mini} & \textbf{Gemini-3-Flash} \\
\midrule
MAP & 0.3078 & 0.3703 & 0.3810 & 0.3793 & 0.3680 & 0.3847 & \textbf{0.4342} \\
MRR & 0.3409 & 0.4014 & 0.4082 & 0.3991 & 0.3991 & 0.4237 & \textbf{0.4575} \\
Recall@5 & 0.4184 & 0.4306 & 0.4618 & 0.4387 & 0.4216 & 0.4787 & \textbf{0.4978} \\
Recall@10 & 0.4705 & 0.5501 & 0.5677 & 0.5099 & 0.5379 & 0.5030 & \textbf{0.5739} \\
\bottomrule
\end{tabular}
\end{adjustbox}
\caption{LOCOMO Category 3: Multi-hop queries (n=96).}
\end{table}

\subsection{Category 4: Open-Domain (n=841)}

\begin{table}[H]
\centering
\small
\begin{adjustbox}{max width=\linewidth}
\begin{tabular}{lccccccc}
\toprule
\textbf{LOCOMO} & \textbf{BGE-v2-m3} & \textbf{MemReranker-0.6B} & \textbf{MemReranker-4B} & \textbf{Qwen3-R-4B} & \textbf{Qwen3-R-8B} & \textbf{GPT-4o-mini} & \textbf{Gemini-3-Flash} \\
\midrule
MAP & 0.7267 & 0.7728 & 0.7829 & 0.7415 & 0.7824 & 0.7787 & \textbf{0.8280} \\
MRR & 0.7323 & 0.7770 & 0.7853 & 0.7456 & 0.7864 & 0.7812 & \textbf{0.8296} \\
Recall@5 & 0.8430 & 0.8775 & 0.8853 & 0.8692 & 0.8882 & 0.8888 & \textbf{0.9055} \\
Recall@10 & 0.8906 & 0.9031 & 0.9150 & 0.9025 & 0.9096 & 0.9037 & \textbf{0.9191} \\
\bottomrule
\end{tabular}
\end{adjustbox}
\caption{LOCOMO Category 4: Open-domain queries (n=841).}
\end{table}

\end{document}